\documentclass[10pt,twocolumn,letterpaper]{article}

\usepackage[pagenumbers]{cvpr}
\usepackage{booktabs}
\usepackage{siunitx}
\usepackage[table]{xcolor}
\definecolor{cvprblue}{rgb}{0.21,0.49,0.74}
\usepackage[breaklinks,colorlinks,allcolors=cvprblue]{hyperref}

\def\confName{CVPR}
\def\confYear{2026}

\title{SignMimic: Robust High-Quality Sign Language Motion Generation via Human-Shape-Oblivious Pose Transfer Guidance}

\author{
Zhewen He$^{1}$ \quad
Junyi Yu$^{1}$ \quad
Haomian Huang$^{1}$ \quad
Zhenhua Li$^{2}$ \quad
Yi Fang$^{1,2}$\thanks{Corresponding author.}\\
$^{1}$New York University Abu Dhabi \qquad
$^{2}$ChatSign Technology
}

\begin{document}
\maketitle


\begin{abstract}
We study the challenge of \emph{sign language video mimicking}: given a driving video and a single reference frame, synthesize a video where the target signer reproduces the source motion while preserving identity and linguistic form. Prior pipelines entangle rigid motion, non-rigid deformation, and view-dependent completion in a monolithic generator, causing handshape drift and spatio-temporal instability. We present \textbf{SignMimic}, which (i) applies a TNet-based Model to study SE(3) rigid canonicalization to stabilize global pose, (ii) performs non-rigid adaptation in a canonical space to preserve fine-grained articulators (hands/face) and coarticulation via NIF2d, and (iii) uses Pose-MAE-style completion before conditional video diffusion. This factorization injects geometric and linguistic priors, yielding \emph{shape} and \emph{spatio-temporal} consistency. On several large-scale datasets (ASL 50K, How2Sign, CSL News), SignMimic achieves SOTA-level on video quality, identity similarity and frame continuity while also achieving minimal loss when performing back translation (SLT) on generated videos. Ablations (–rigid/–non-rigid/–completion) confirm each stage’s role. Code is available at \url{https://anonymous.4open.science/r/UniSignMimicTurbo-6088}; model checkpoints and video examples will be released.
\end{abstract}
\section{Introduction}
\label{sec:intro}
Signer-specific \emph{sign language video mimicking} aims to render a target signer reproducing the motion of a driving video while preserving identity and linguistic form (handshape, location, movement, and non-manual signals, NMS). It empowers content creation, data augmentation for SLT/ASR, and accessibility tools, extending to other motion mimicking in video generation tasks. 

\subsection{The Challenge of Sign Video Mimicking}
\label{sec:intro_challenge}
While many video pose mimicking pipelines  \cite{Zhang2024MimicMotion, MooreThreadsAnimateAnyone, Chang2024MagicPose}, \emph{sign language} mimicking is particularly demanding as a sub-question of video mimicking: it requires precise articulator motion (hands/face) and strong spatio-temporal continuity for linguistic intelligibility. This makes the problem both challenging and impactful.

We observe that existing pipelines often \textbf{entangle} three distinct factors within
a single generator: (1) \emph{rigid motion} (global $SE(3)$ pose of the body/skeleton),
(2) \emph{non-rigid deformation} (articulated hand motion, facial expressions, clothing),
and (3) \emph{view-dependent completion} (disocclusions and missing details).
Such entanglement commonly yields three failure modes: (a) \textit{shape inconsistency}
(e.g., handshape drift, finger collapse), (b) \textit{spatial inconsistency}
(e.g., limb misalignment, self-intersections), and (c) \textit{temporal inconsistency}
(e.g., jitter and drift)—all detrimental to linguistic accuracy and comprehension.

\subsection{Our Insight}
\label{sec:intro_insight}
We argue that achieving \emph{shape} and \emph{spatio-temporal} consistency requires
\textbf{disentangling conditioning} before video synthesis. Instead of asking a
monolithic model to learn everything at once, we canonicalize, adapt, and complete:
(i) \emph{SE(3) rigid canonicalization} stabilizes global pose in a shared canonical
space; (ii) a \emph{non-rigid deformation field} (NIF2D) adapts canonical features
to the target signer, preserving fine articulators and coarticulation; (iii)
\emph{pose-aware masked completion} (Pose MAE) fills pose-induced disocclusions,
providing clean, geometry-aware inputs to a conditional video diffusion model.
This factorization injects geometric and linguistic priors and offloads temporal
stability from the generator to the conditioning pipeline.

\subsection{SignMimic in brief}
\label{sec:intro_unisign}
\textbf{SignMimic} instantiates the above principle with a T-Net for per-frame
SE(3) canonicalization, a 2D neural deformation field aligned in the canonical space,
and a masked completion head that uses pose-derived visibility cues. The diffusion
backend then focuses on photorealism and local detail while inheriting stable,
geometry-consistent conditioning. The design is modular, data-agnostic, and compatible
with common detectors/encoders for poses and identity. Following listed is our main contributions:

\subsection{Contributions}
Our main contributions are four-fold. (1) We propose \textbf{SignMimic}, a modular pipeline that disentangles sign language mimicking into rigid canonicalization, non-rigid adaptation, and pose completion to improve consistency. (2) We introduce geometric and temporal regularizers (e.g., SE(3) smoothness) to curb shape drift and stabilize coarticulation. (3) We establish a comprehensive evaluation protocol covering video quality, frame continuity, identity fidelity, and linguistic accuracy via back-translation. (4) We demonstrate state-of-the-art results on four large-scale datasets, confirming the effectiveness of our factorized conditioning approach.

\section{Related Work}
\label{sec:related_work}

\subsection{Human Pose Extraction}
\label{sec:related_work_pose}
2D/3D human pose estimation has progressed rapidly, from bottom-up keypoint grouping
to top-down detectors with high-resolution backbones \cite{cao2019openpose,sun2019deep,sun2021monocular,lugaresi2019mediapipe}.
For sign language, hand- and face-centric estimators are crucial: accurate handshape,
location, and facial landmarks (NMS) strongly correlate with intelligibility
\cite{lugaresi2019mediapipe,simon2017hand,bulat2017far}. In practice, pose extractors trade
off robustness and granularity: whole-body models provide consistent skeletons in
the wild, while specialized hand/face heads improve local precision under occlusion.
SignMimic is agnostic to the specific extractor; we use off-the-shelf keypoints
and confidence maps to drive canonicalization and to build pose-aware masks.

\subsection{Rigid \& Non-Rigid Transformation}
\label{sec:related_work_transform}
Spatial Transformer Networks introduced differentiable image warping and coarse
canonicalization \cite{jaderberg2015spatial}. Subsequent work learned canonical spaces and deformation
fields for articulated or dynamic scenes, in both 2D and 3D (e.g., canonical NeRFs
with per-frame warps) \cite{martin2021nerf,park2021nerfies,li2021neural}. For mimicking tasks,
rigid SE(3) motion stabilizes global pose, while non-rigid fields capture joint-level
articulation and clothing deformations. Our design follows this separation: a T-Net
estimates per-frame rigid transforms into a shared canonical space, and a lightweight
2D deformation field (NIF2D) adapts canonical features to signer-specific geometry,
regularized by SE(3) smoothness and Jacobian constraints.

\subsection{MAE Networks}
\label{sec:related_work_mae}
\subsubsection{Image MAE} Masked autoencoders (MAE) reconstruct missing content from sparsely observed tokens, providing strong priors for inpainting and representation learning \cite{He2022MAE,bao2021beit}. Variants have been explored for images and videos, with pose- or visibility-guided masking improving structure recovery in articulated motion \cite{tong2022videomae,deng2019deep}.

\subsubsection{Skeleton MAE} Beyond pixels, MAE has been adapted to \emph{skeleton sequences} by masking joints and/or temporal windows and reconstructing kinematic signals \cite{Wu2022SkeletonMAE,jiang2020maskgcn,li2023aim_mae}. These methods typically operate on graph-structured inputs with joint/bone encodings, and incorporate objectives on coordinates, velocities, and bone-length consistency to inject strong kinematic priors. Skeleton-level masked modeling has improved downstream
tasks such as action recognition, pose forecasting, and motion synthesis, and is complementary to image/video MAE.

\begin{figure*}[t]
    \centering 
    \includegraphics[width=\textwidth]{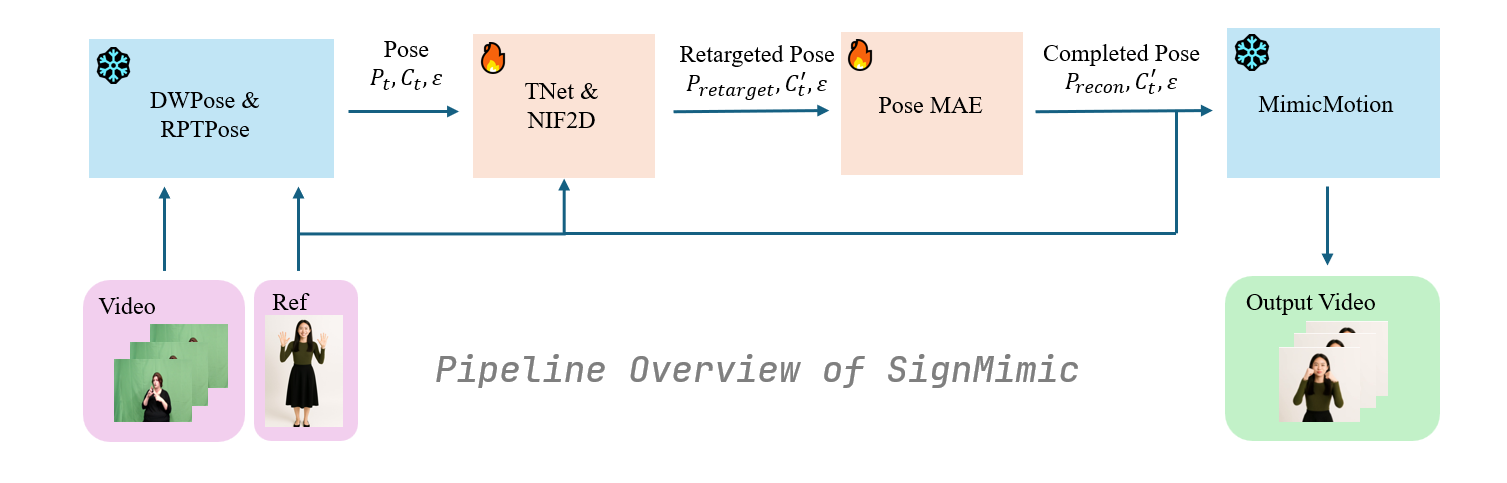}
    \caption{
        \textbf{An overview of the SignMimic pipeline.}
        Our framework first extracts a pose sequence from a source video via DWPose.
        This pose is then adapted to a reference identity by our core retargeting modules (TNet \& NIF2D) and refined by a Pose MAE to handle occlusions.
        Finally, a pre-trained generator (MimicMotion) synthesizes the output video.
        Blue (snowflake icon) indicates frozen, pre-trained models, while orange (flame icon) indicates trainable modules.
    }
    \label{fig:pipeline}
\end{figure*}

\subsection{Video Diffusion Models for Pose Mimicking}
\label{sec:related_diffusion}
\subsubsection{Diffusion Models for Video Synthesis}
Denoising diffusion models, from score-based SDEs to the now-prevalent Latent Diffusion (LDM), have become the state-of-the-art for high-fidelity synthesis due to their strong performance and flexible conditioning via classifier-free guidance \cite{Ho2020DDPM, song2020score_sde, Rombach2022LDM, ho2022classifier}. This paradigm was extended to video with models like Stable Video Diffusion (SVD), which incorporate temporal attention mechanisms to maintain coherence across frames \cite{blattmann2023svd}. However, a fundamental challenge persists for mimicking tasks: naively applying these powerful generators entangles global motion, local deformation, and completion, which can cause temporal drift and jitter. This issue is exacerbated by their high sensitivity to the quality of conditioning signals, as any errors or occlusions in an input pose sequence will invariably propagate through the generator and degrade the final output.

\subsubsection{Controlled SVD for Motion Mimicking Tasks}
To mimic motion, recent work augments SVD with structured controls—keypoints,
skeletons, optical flow, or depth—via ControlNet-style adapters or cross-attention injections \cite{chen2023controlnet,ma2023pose_diffusion,min2023flow_diffusion}.
These controls help anchor global motion and limb layout, yet monolithic pipelines often mix rigid motion, non-rigid deformation, and inpainting in the diffusion stage, which limits consistency. Our approach keeps diffusion as the \emph{final} renderer: after SE(3) canonicalization, non-rigid adaptation, and pose-aware (Skeleton-/Pose-)MAE completion, the diffusion backend focuses on photorealism while inheriting geometry-
consistent, occlusion-completed conditioning suitable for sign-language mimicking.

\subsection{Sign Language Translation (SLT)}
SLT maps sign videos to text using CNN/Transformer pipelines, sometimes with gloss as an intermediate representation \cite{Tarres2023How2SignSLT}. Benchmarks such as How2Sign provide standard splits and automatic metrics (BLEU, chrF++) \cite{Duarte2021How2Sign, Papineni2002BLEU}. For mimicking, back-translation serves as a proxy for linguistic fidelity: lower SLT loss due to mimicking on generated videos implies better preservation of phonological form and NMS coherence. We therefore report the difference of BLEU-1/2/3/4, rBLEU, and chrF++ between SignMimic and our benchmark (\S\ref{sec:exp_sota}).

\section{Method}
\label{sec:method}

\subsection{Human Pose Extraction}
\label{sec:method_pose}
We mainly adopt \textbf{DWPose} as the 2D whole-body keypoint extractor for each frame
$I_t$ and \textbf{RTMPose} as fallback or a quick solution \cite{Yang2023DWPose, Jiang2023RTMPose}. Those extractor can both return COCO-Whole-Body-like four parts of keypoints and confidences:
body, face, right hand, and left hand \cite{Jin2020COCOWholeBody}. We denote the per-frame joint coordinates
and confidences as
\[
P_t \in \mathbb{R}^{J\times 2},\quad
C_t \in [0,1]^J,
\]
where $J=J_{\text{body}}+J_{\text{face}}+J_{\text{rh}}+J_{\text{lh}}$ and the four
subsets are indexed by $\mathcal{J}_{\text{body}},\mathcal{J}_{\text{face}},
\mathcal{J}_{\text{rh}},\mathcal{J}_{\text{lh}}$. DWPose also provides a
\emph{subset} of edges (bone links) $\mathcal{E}\subseteq \{(i,k)\mid i,k\in [1..J]\}$,
which we use as the kinematic graph. We write the whole-body skeleton as
$\mathcal{G}=(\mathcal{V},\mathcal{E})$ with $\mathcal{V}=[1..J]$.

\subsection{Rooting and scale normalization.}
To achieve invariance to translation and scale, we normalize the keypoints of each part $\mathcal{J}_{\text{part}} \in \{\mathcal{J}_{\text{body}}, \mathcal{J}_{\text{face}}, \mathcal{J}_{\text{rh}}, \mathcal{J}_{\text{lh}}\}$ independently. For a given part, we compute the bounding box enclosing all its keypoints $\{p_i = (x_i, y_i)\}_{i \in \mathcal{J}_{\text{part}}}$. Let the center of this bounding box be $(c_x, c_y)$ and its dimensions be $(w, h)$. Each keypoint $p_i$ is then normalized to $\hat{p}_i = (\hat{x}_i, \hat{y}_i)$ using the following transformation:
\begin{equation}
\hat{x}_i = \frac{x_i - c_x}{w}, \quad \hat{y}_i = \frac{y_i - c_y}{h}.
\label{eq:pose_normalization}
\end{equation}
This per-part normalization preserves the internal geometric structure of each component (e.g., hand shape) while canonicalizing its global position and scale.

\subsection{TNet in Rigid Transformation}
\label{sec:method_tnet}

\subsubsection{Rigid Canonicalization via T-Net}
To absorb the \emph{rigid} misalignments between the source and target signers (e.g., global scale, part-specific offsets), which are a major source of identity drift, we repurpose the canonicalization idea from PointNet's T-Net~\cite{Qi2017PointNet}. We treat the 2D keypoints of each body part (body, face, and hands) as a small, unordered point set. A lightweight, part-specific T-Net then regresses a 2D affine transformation to align each part. This data-driven approach effectively normalizes scale and offset differences in a permutation-invariant manner, providing a clean, rigidly-aligned pose for the subsequent non-rigid adaptation stage (\S\ref{sec:method_nif2d}).

\subsubsection{Architecture}
We build a \emph{Hierarchical Pose Grafting Network} with three expert encoders:
body, face, and a shared hand encoder. Each encoder begins with a TNet that predicts
a $2{\times}2$ alignment matrix $A_p$ for part $p$ and applies a pointwise transform
$x\!\mapsto\!xA_p$; shared $1{\times}1$ convolutions and global max-pooling then yield
a part descriptor $g_p$. Given a \emph{reference} pose and a \emph{driving} pose, we
encode four parts for each, concatenate the eight descriptors, and use an MLP decoder
to predict the grafted whole-body keypoints.

\begin{table*}[t]
    \centering
    \small
    \setlength{\tabcolsep}{4pt}
    \renewcommand{\arraystretch}{1.05}
    \begin{tabular}{p{3.0cm} p{4.0cm} p{8.5cm}}
    \toprule
    \textbf{Loss} & \textbf{Applies to} & \textbf{Definition} \\
    \midrule
    \textbf{Masked reconstruction} $\mathcal{L}_{\text{recon}}$
    & All parts; masked \& high-conf joints
    & $\displaystyle \mathcal{L}_{\text{recon}}=\frac{1}{|\Omega|}\sum_{(t,j)\in\Omega}\rho_\delta\!\big(\hat X_{t,j}-X_{t,j}\big)$ \\[4pt]
    
    \textbf{Velocity} $\mathcal{L}_{\text{vel}}$
    & All parts; masked joints
    & $\displaystyle \mathcal{L}_{\text{vel}}=\frac{1}{|\Omega_v|}\sum_{(t,j)\in\Omega_v}\!\big\|\Delta\hat X_{t,j}-\Delta X_{t,j}\big\|_2^{2}$ \\[4pt]
    
    \textbf{Acceleration} $\mathcal{L}_{\text{acc}}$
    & All parts; masked joints
    & $\displaystyle \mathcal{L}_{\text{acc}}=\frac{1}{|\Omega_a|}\sum_{(t,j)\in\Omega_a}\!\big\|\Delta^{2}\hat X_{t,j}-\Delta^{2}X_{t,j}\big\|_2^{2}$ \\[4pt]
    
    \textbf{Total variation} $\mathcal{L}_{\text{tv}}$
    & All parts; masked joints (predictions only)
    & $\displaystyle \mathcal{L}_{\text{tv}}=\frac{1}{|\Omega_v|}\sum_{(t,j)\in\Omega_v}\!\big\|\Delta\hat X_{t,j}\big\|_{1}$ \\[4pt]
    
    \textbf{Body bone length} $\mathcal{L}_{\text{bone}}$
    & Body stream; valid bones by \texttt{subset}
    & $\displaystyle \mathcal{L}_{\text{bone}}=\frac{1}{\sum_t|\mathcal{B}^{\text{valid}}_t|}\!
    \sum_{t}\!\sum_{(i,k)\in\mathcal{B}^{\text{valid}}_t}
    \Big(\,\|\hat X_{t,i}-\hat X_{t,k}\|_2-\|X_{t,i}-X_{t,k}\|_2\Big)^{2}$ \\
    \bottomrule
    \end{tabular}
    \vspace{4pt}
    \caption{\textbf{Pose MAE Losses}
    $X,\hat X\in\mathbb{R}^{T'\times V\times 2}$; $\Delta X_{t,j}=X_{t,j}-X_{t-1,j}$; $\Delta^2$ is the second difference.
    $\Omega$ is the set of masked \& high-confidence joints; $\Omega_v{=}\{(t\!\ge\!2,j)\!\in\!\Omega\}$; $\Omega_a{=}\{(t\!\ge\!3,j)\!\in\!\Omega\}$.
    $\mathcal{B}^{\text{valid}}_t$ are body limb pairs with both endpoints present per \texttt{subset}.}
    \label{tab:mae_losses}
\end{table*}

\subsection{NIF2D in Non-rigid Transform}
\label{sec:method_nif2d}

\subsubsection{Problem and Scope}
While the TNet-based rigid canonicalization (§\ref{sec:method_tnet}) successfully aligns the global pose and corrects for part-scale mismatches, it is by design incapable of modeling non-rigid deformations. These are critical for sign language, encompassing fine-grained hand articulations, facial expressions (Non-Manual Signals), and coarticulation effects that are specific to the target signer's morphology. 

\subsubsection{Method}
To this end, we introduce a 2D Neural Implicit Deformation Field (NIF2D). To ensure the model is compatible with arbitrary reference identities, we first employ a dedicated \textbf{Identity Encoder}, $E_{\text{id}}$, to produce a generalized identity embedding. This encoder takes the keypoints of the reference signer's pose, $P_{\text{ref}}$, and maps them to a compact latent vector $\mathbf{z}_{\text{ref}} = E_{\text{id}}(P_{\text{ref}})$. This vector is trained to capture pose-invariant characteristics, such as limb proportions and morphological style, focusing primarily on the fine-grained articulators (hands and face) which contain the most salient geometric information.

The NIF2D, modeled as a lightweight MLP $f_{\theta}$, is then conditioned on this dynamic identity vector $\mathbf{z}_{\text{ref}}$ to predict the necessary per-keypoint offset $\Delta p_i$ for a given keypoint $p_i^{\text{canon}}$ from the rigidly-aligned driving pose:
\begin{equation}
    \Delta p_i = f_{\theta}(p_i^{\text{canon}}, \mathbf{z}_{\text{ref}}),
\end{equation}
where $\mathbf{z}_{\text{ref}} = E_{\text{id}}(P_{\text{ref}})$ is the embedding produced by the identity encoder. The final retargeted keypoint is then computed as $p_i^{\text{retarget}} = p_i^{\text{canon}} + \Delta p_i$. This design decouples identity encoding from the deformation process, allowing our model to generalize to unseen subjects at inference time by simply providing a new reference image.

\subsection{Transformer-based Pose MAE}
\label{sec:method_posemae}
We adopt an \textbf{asymmetric encoder--decoder} masked autoencoding scheme on
pose sequences inspired by Classic MAE and Skeleton MAE \cite{He2022MAE, Wu2022SkeletonMAE}. Given per-frame keypoints $P_t$ (Sec.~\ref{sec:method_pose}) and the
kinematic subset $\mathcal{E}$, the MAE receives a part-aware sequence
(body / face / left-hand / right-hand) and reconstructs masked joints over time.

\subsubsection{Masking strategies.}
We use three complementary masks to simulate realistic missingness:
(i) \emph{confidence mask} that drops low-quality joints using a threshold
$\tau_{conf}$ on detector confidence; (ii) \emph{temporal mask} that removes a random
$x\%$ of frames per clip; and (iii) \emph{spatial mask} that removes a random
subset of keypoints within a frame. 

\subsubsection{Pose MAE Architecture}
Firstly, we tokenize \emph{per part} (body/face/left hand/right hand) at the \emph{per-keypoint} level—
each token embeds $(x,y)$, confidence, and a part ID. This localizes modeling to
semantically coherent joints, providing a useful inductive bias and cutting the
attention space from $\mathcal{O}(V^2)$ to $\mathcal{O}\!\big(\sum_p V_p^2\big)$. This part-wise design, inspired by UniSign \cite{Li2025UniSign}, is quite noteworthy because we witnessed that in practice it converges faster and stabilizes hand/face
reconstruction.

Then, a lightweight 1D temporal conv stack (shared within each part)
extracts short-range dynamics before a cross-part Transformer encoder aggregates
global context. 

The encoder only processes \emph{visible} tokens; masked tokens
are replaced by learned mask embeddings and are handled solely by a small decoder.
We add (i) \textit{joint and temporal positional embeddings}, (ii) \textit{part
embeddings} to preserve hand/face/body semantics, and (iii) an optional \textit{
graph bias} from the kinematic subset $\mathcal{E}$ (used as an attention prior
on the body stream). The decoder attends to encoder outputs plus mask embeddings
and predicts the missing 2D coordinates for masked joints; unmasked joints are
passed through.

\subsubsection{Self-supervised objectives.}
Losses are computed \emph{only on masked and high-confidence} joints and we applied these following losses for training our Pose MAE:
\begin{itemize}
  \item \textbf{Masked reconstruction} $\mathcal{L}_{\text{recon}}$:
        a per-part Huber loss between predictions and ground truth.
  \item \textbf{Velocity consistency} $\mathcal{L}_{\text{vel}}$:
        encourages first-order temporal consistency on masked joints.
  \item \textbf{Acceleration consistency} $\mathcal{L}_{\text{acc}}$:
        penalizes second-order temporal deviations for smooth coarticulation.
  \item \textbf{Total variation} $\mathcal{L}_{\text{tv}}$:
        regularizes high-frequency jitter along time for masked joints.
  \item \textbf{Body bone-length} $\mathcal{L}_{\text{bone}}$:
        preserves limb lengths on the body graph using the provided subset
        connectivity; applied to the body stream only.
\end{itemize}
The total loss is a weighted sum
$\mathcal{L}_{\text{MAE}}=
\lambda_{\text{recon}}\mathcal{L}_{\text{recon}}
+\lambda_{\text{vel}}\mathcal{L}_{\text{vel}}
+\lambda_{\text{acc}}\mathcal{L}_{\text{acc}}
+\lambda_{\text{tv}}\mathcal{L}_{\text{tv}}
+\lambda_{\text{bone}}\mathcal{L}_{\text{bone}}$. The detailed formula and weights are shown in Table \ref{tab:mae_losses}.

\subsection{Video Mimicking}
\label{sec:method_vid}
We employ diffusion strictly as the \emph{final renderer}. Upstream conditioning
(rigid TNet canonicalization, NIF2D, and Pose MAE completion) produces
geometry-consistent and occlusion-completed signals; the diffusion stage only
performs video synthesis at inference time.

\subsubsection{Pretrained backbones}
We \textbf{reuse} public pretrained implementations:
(i) \textbf{MimicMotion} as a pose-conditioned video generator \cite{Zhang2024MimicMotion}; and
(ii) \textbf{Diffusers} to run latent video diffusion models (e.g.,
SVD-style) when needed \cite{von_platen_patrick_2022_diffuser}. 

\subsubsection{Benchmark protocol with MimicMotion}
MimicMotion is used in two roles:
\textbf{(a) Baseline}—run \emph{as-is} on the same keypoints/reference without our
upstream processing; and
\textbf{(b) SignMimic (MM-backend)}—run with \emph{identical pretrained weights}
but driven by our cleaned conditioning (TNet+NIF2D+Pose MAE). This isolates the contribution
of our geometric factorization while holding the diffusion backbone and its parameters
constant.

\section{Experiments}
\label{sec: exp}
We assess \textbf{SignMimic} on four large-scale sign datasets and compare
to the prior state-of-the-art \textit{MimicMotion}~\cite{Zhang2024MimicMotion}.
We focus on: (i) distribution-level video realism, (ii) identity preservation
and temporal continuity, and (iii) linguistic fidelity under SLT back-translation.

\begin{table*}[t]
    \centering
    \small
    \setlength{\tabcolsep}{5pt}
    \renewcommand{\arraystretch}{1.15}
    \begin{tabular}{l l r r r r r r}
    \toprule
    \textbf{DataSet} & \textbf{Method} & \textbf{FID\_VID}~($\downarrow$) & \textbf{FVD}~($\downarrow$) & \textbf{ID\_COS}~($\uparrow$) & \textbf{PSNR\_TS}~($\uparrow$) & \textbf{SSIM\_TS}~($\uparrow$) & \textbf{\boldmath$\Delta$BLEU}~($\uparrow$) \\
    \midrule
    ASL 50K               & SignMimic (Ours)      &  \textbf{55.8082} &  \textbf{471.3722} & 0.4451 & \textbf{28.9048} & 0.9634 &  \textbf{-0.89} \\
    ASL 50K               & SignMimic (w/o MAE)   &  60.5222          &  837.8924          & \textbf{0.4744} & 29.8137 & \textbf{0.9662} &        \\
    ASL 50K               & SignMimic (w/o TNet)  &  97.2169          & 1222.6924          & 0.1451 & 24.6959 & 0.9214 &        \\
    ASL 50K               & SignMimic (w/o NIF2D) &   62.5249                &   678.2377                 &    0.1759    &     25.8771    &    0.9299    &        \\
    ASL 50K               & Mimicmotion (Benchmark)             & 111.3754          & 1309.0642          & 0.1523 & 23.7492 & 0.9038 &  -2.67 \\
    \midrule
    How2Sign              & SignMimic (Ours)      &  \textbf{64.3899} & \textbf{1139.6051} & \textbf{0.3892} & \textbf{25.8220} & \textbf{0.9582} &        \\
    How2Sign              & Mimicmotion  (Benchmark)            &  81.7876          & 1848.9899          & 0.1707 & 23.2193 & 0.9263 &        \\
    \midrule
    CSL News              & SignMimic (Ours)      &  \textbf{52.2940} & \textbf{1261.1849} & \textbf{0.5122} & \textbf{28.8641} & \textbf{0.9670} &        \\
    CSL News              & Mimicmotion (Benchmark)             &  79.0372          & 1771.2363          & 0.2121 & 21.7165 & 0.9049 &        \\
    \midrule
    Phoenix-2014              & SignMimic (Ours)      &  \textbf{68.7545} & \textbf{1027.5737} & \textbf{0.3575} & \textbf{21.9198} & \textbf{0.9312} &        \\
    Phoenix-2014              & Mimicmotion (Benchmark)             &  90.0887          & 1514.9813          & 0.1885 & 19.8100 & 0.8899 &        \\
    \bottomrule
    \end{tabular}
    \caption{\textbf{Quantitative results.} We evaluate SignMimic against the previous SOTA model MimicMotion on four large-scale datasets: ASL 50K, How2Sign, CSL News and Phoenix-2014. Ablations on TNet, NIF2D, and PoseMAE confirm our contributions. Arrows indicate better directions, data with bold font indicates the best result.}
    \label{tab:main_results}
\end{table*}

\begin{figure*}[t]
    \centering
    \includegraphics[width=\textwidth, height=6cm]{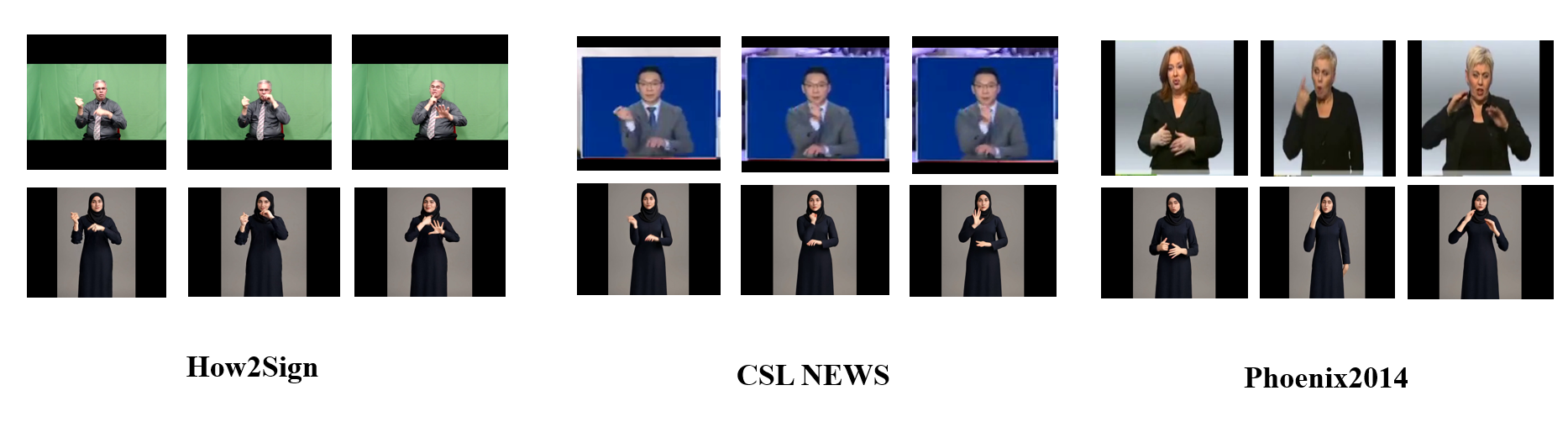}
    \caption{\textbf{Qualitative results}. The top row displays original video frames from three datasets: How2Sign, CSL News, and Phoenix-2014.  The bottom row shows the corresponding mimicked results. Notably, all mimicked images were generated using \textbf{only a single} reference image. (Raw images from ASL 50K Dataset is not available for publication use due to \textbf{copyright} issues)}
    \label{fig:demo}
\end{figure*}

\begin{figure}[t]
    \centering 
    \includegraphics[width=0.5\textwidth, height=8cm]{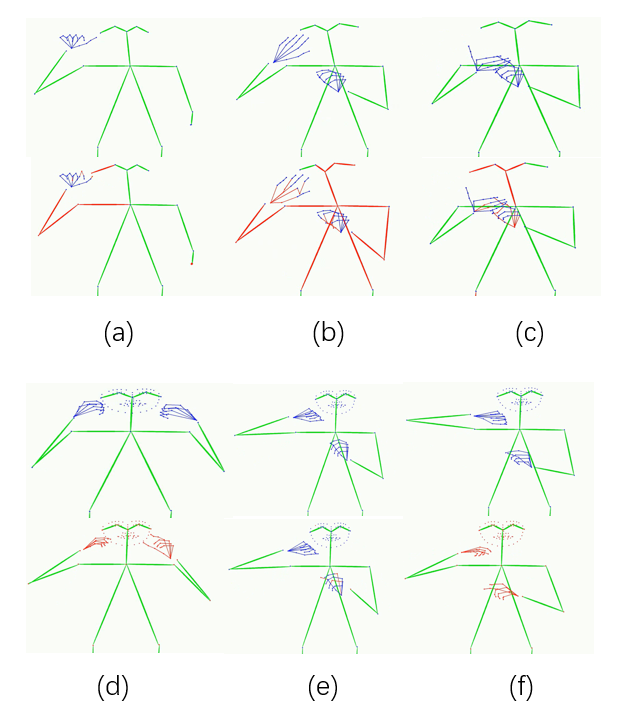}
    \caption{\textbf{Qualitative results of PoseMAE}. This figures visualizing the ability of our \textbf{PoseMAE}. 1) Part-wise attention: (a) - (c) shows completion of body and hands, while (d) - (f) shows completion of hands and faces. 2) Different masking strategies, (a),(c),(e) shows spatial masking, while (b), (d), (f) shows temporal masking.}
    \label{fig:mae_pose}
\end{figure}

\subsection{Experiment Setup}
\label{sec:exp_setup}
\subsubsection{Datasets.}
We evaluate on \textbf{ASL~50K}, \textbf{How2Sign} (front-facing test split)
~\cite{Duarte2021How2Sign}, \textbf{CSL~News} (released with Uni-Sign) and \text{Phoenix-2014} \cite{CSLNews2025, Camgoz2018PHOENIX2014T}. Whole-body 2D poses (17 body, 6 feet, 68 face, 42 hands;
total 133) follow the COCO-WholeBody topology~\cite{Jin2020COCOWholeBody}. Keypoints
are extracted by \textbf{DWPose}~\cite{Yang2023DWPose}.

\subsubsection{Pipeline Inference Steps}
Our inference pipeline is a fixed, sequential process. For a given driving video and reference frame, we first extract and normalize 2D whole-body keypoints using DWPose. These are then processed sequentially by our core geometric modules: the TNet for rigid canonicalization, NIF2D for non-rigid adaptation, and the Pose-MAE for completion and smoothing. The final, cleaned pose sequence is then de-normalized and fed into the pre-trained MimicMotion renderer to synthesize the output video. Further implementation details are provided in the  Appendix (a.Training \& Inference Details).

\subsubsection{Metrics and protocol.}
We report (1) \emph{Video quality:} FID(Video Version) and FVD~\cite{Heusel2017FID,Unterthiner2018FVD};
(2) \emph{Frame continuity:} PSNR-TS and SSIM-TS (sequence-wise averages);
(3) \emph{Mimicking/identity:} face-embedding cosine similarity using
InsightFace Buffalo~\cite{InsightFaceBuffaloL,Deng2019ArcFace}; and
(4) \emph{SLT fidelity:} $\Delta$BLEU$_4$ (BLEU$_4$ on generated videos minus BLEU$_4$
on real videos using a frozen SLT model; less negative is better)~\cite{Papineni2002BLEU,Post2018SacreBLEU,PopovicChrf}.
Lower is better for FID/FVD; higher is better otherwise. For FVD/FID we follow
standard I3D features and protocol in prior work; all methods share the same
test splits and evaluation code.

\subsection{Comparison with Previous SOTA Models}
\label{sec:exp_sota}
Table ~\ref{tab:main_results} shows consistent gains over \textit{MimicMotion}
on all three datasets \cite{Zhang2024MimicMotion}. On \textbf{ASL~50K}, FID\_VID and FVD drop from
$111.38 \rightarrow \mathbf{55.81}$ and $1309.06 \rightarrow \mathbf{471.37}$,
respectively, while ID-COS improves from $0.152 \rightarrow \mathbf{0.445}$ and
PSNR/SSIM increase by $+5.16$ dB / $+0.0596$. The SLT degradation is also reduced
($\Delta$BLEU$_4$: $-2.67 \rightarrow \mathbf{-0.89}$), indicating stronger
preservation of linguistic content. Similar trends hold on \textbf{How2Sign},
\textbf{CSL~News} and \textbf{Phoenix-2014}, where SignMimic attains the best scores across video quality,
identity, and temporal metrics. These improvements suggest that factorized
conditioning (rigid/non-rigid/completion) benefits diffusion-based mimicking beyond
simply copying per-frame appearance.


\definecolor{spatialcolor}{rgb}{0.12, 0.35, 0.57} 
\definecolor{temporalcolor}{rgb}{0.0, 0.5, 0.25}  
\definecolor{confcolor}{rgb}{0.7, 0.25, 0.15}   
\begin{table*}[t]
\centering
\label{tab:ablation_study}
\sisetup{tight-spacing = true} 
\begin{tabular}{
    c c c 
    S[table-format=1.2e-1]
    S[table-format=1.2e-1]
    S[table-format=1.2e-1]
    S[table-format=1.2e-1]
    S[table-format=1.2e-1]
    S[table-format=1.2e-1]
}
\toprule
\multicolumn{3}{c}{\textbf{Mask Parameters}} & \multicolumn{6}{c}{\textbf{Average Loss Components}} \\ 
\cmidrule(lr){1-3} \cmidrule(lr){4-9}
{\textbf{Spatial}} & {\textbf{Temporal}} & {\textbf{Conf}} & {\textbf{Accel.}} & {\textbf{Bone}} & {\textbf{Recon.}} & {\textbf{Total}} & {\textbf{TV}} & {\textbf{Velocity}} \\ 
\midrule

\textbf{15} & \textbf{5} & \textbf{3} & \bfseries 6.09e-4 & \bfseries 2.09e-4 & \bfseries 1.78e-4 & \bfseries 8.87e-4 & \bfseries 6.17e-2 & \bfseries 2.11e-4 \\
\midrule

\multicolumn{9}{l}{\textit{\textcolor{spatialcolor}{Experiment on Spatial Mask}}} \\
\textbf{5}  & 5 & 3 & 3.62e-4 & 1.80e-4 & 1.46e-4 & 5.75e-4 & 3.71e-2 & 1.24e-4 \\
\textbf{25} & 5 & 3 & 7.55e-4 & 2.13e-4 & 1.79e-4 & 1.08e-3 & 7.91e-2 & 2.62e-4 \\
\midrule

\multicolumn{9}{l}{\textit{\textcolor{temporalcolor}{Experiment on Temporal Mask}}} \\
15 & \textbf{10} & 3 & 7.25e-4 & 2.09e-4 & 1.84e-4 & 1.03e-3 & 7.40e-2 & 2.51e-4 \\
15 & \textbf{15} & 3 & 7.20e-4 & 2.09e-4 & 1.83e-4 & 1.03e-3 & 7.40e-2 & 2.49e-4 \\
\midrule

\multicolumn{9}{l}{\textit{\textcolor{confcolor}{Experiment on Confidence Mask}}} \\
15 & 5 & \textbf{5} & 8.96e-4 & 2.04e-4 & 2.27e-4 & 9.84e-4 & 6.26e-2 & 3.13e-4 \\
15 & 5 & \textbf{7} & 1.10e-3 & 2.13e-4 & 2.25e-4 & 1.04e-3 & 6.54e-2 & 3.89e-4 \\
\bottomrule
\end{tabular}
\caption{
    \textbf{Quantitive study of different masking settings of PoseMAE}
    This table analyzes the impact of the spatial mask ratio, temporal mask ratio, and confidence mask threshold settings on model performance, measured by various loss components. 
    Our baseline configuration is (Spatial=15, Temporal=5, Conf=3), highlighted in bold. 
    Each subsequent section ablates a single parameter while holding the others constant to isolate its effect. All loss values are presented in scientific notation.
}
\end{table*}
\subsection{Ablation Study}
\label{sec:exp_ablation}
We isolate each component on \textbf{ASL~50K}:
\begin{itemize}
  \item \textbf{w/o TNet} \textemdash{} removing rigid canonicalization severely hurts realism
        and identity (FID/FVD: $55.81/471.37 \rightarrow 97.22/1222.69$; ID-COS:
        $0.445 \rightarrow 0.145$), underscoring the need to normalize global scale
        and offsets before diffusion.
  \item \textbf{w/o MAE} \textemdash{}  dropping pose completion increases FID/FVD
        ($55.81/471.37 \rightarrow 60.52/837.89$) despite slightly higher
        ID-COS/PSNR/SSIM. This indicates our MAE acts as a structure-aware
        regularizer that improves distribution-level realism and spatio-temporal
        stability rather than overfitting per-frame details.
  \item \textbf{w/o NIF2D} \textemdash{} removing non-rigid adaptation degrades realism
        and identity (FID/FVD: $55.81/471.37 \rightarrow 62.52/678.24$; ID-COS:
        $0.445 \rightarrow 0.176$), confirming its role in preserving articulators
        and coarticulation.
\end{itemize}

\subsection{Extended Study with Pose MAE}
\label{sec:exp_mae}
\subsubsection{Masking Parameters Experiment}
We study three complementary masks in the MAE inference path:
(i) \textbf{confidence mask} $M_c$ that drops joints whose detector confidence
is below a threshold $\tau$;
(ii) \textbf{temporal mask} $M_t$ that removes a fraction $\rho_t$ of frames
(regular or random stride) to test temporal completion; and
(iii) \textbf{spatial mask} $M_s$ that hides a fraction $\rho_s$ of keypoints per
frame. The effective mask is $M = M_c \lor M_t \lor M_s$.
We sweep $\tau \in \{0.3,0.5,0.7\}$, $\rho_t \in \{0.05,0.10,0.15\}$,
$\rho_s \in \{0.05,0.15,0.25\}$ and report per-part errors (body/face/left hand/right hand)
with the metrics defined in Sec.~\ref{sec:method_posemae}:
masked \textit{Huber} $\mathcal{L}_{\text{recon}}$, velocity/acceleration
consistency, TV, and the body bone-length loss. 
\subsubsection{MAE Experiment Result}
The quantitative results are presented in Table~\ref{tab:mae_losses}. We observe that increasing the masking ratios---whether spatial (up to 25\%) or temporal (up to 15 frames)---consistently leads to higher reconstruction and motion-related losses (velocity and acceleration). This indicates that while masking is essential for representation learning, excessive sparsity challenges the recovery of fine-grained dynamics. Our baseline setting (Spatial=15, Temporal=5) strikes an optimal balance between difficulty and stability. Qualitatively, Figure~\ref{fig:mae_pose} further validates these findings. The visualizations demonstrate that our PoseMAE, driven by part-wise attention, effectively hallucinates plausible geometries for missing hands and faces under both spatial occlusion and temporal dropout scenarios, ensuring structurally complete inputs for the downstream diffusion process.

\subsubsection{Application: Sentence-level Interpolation}
To generate continuous, sentence-level sign language videos from our dataset of isolated word-level clips, a key challenge is creating natural transitions between them. We address this by introducing a novel transition smoothing method. First, we leverage our Pose MAE for temporal tracking by treating the gap between two clips as a mask to be inpainted. The MAE predicts plausible intermediate poses based on the context of the preceding and succeeding clip. 
Subsequently, these predictions guide a lightweight Spherical Linear Interpolation (\textit{slerp}) to generate a perfectly smooth motion path. Related Experiments is displayed in Appendix (b. Interpolation Experiment) due to the page limitation.

\subsection{Back Translation}
\label{sec:back_translation}
\subsubsection{Experiment Setup}
To quantify linguistic fidelity, we mainly perform \emph{back translation} on
\textbf{How2Sign} \cite{Duarte2021How2Sign}. For each test segment we have (i) the ground-truth text
reference and (ii) a generated video from our pipeline or from the baseline.
We evaluate with the official \textbf{How2Sign SLT} model provided by the dataset
authors (frozen weights and decoding hyperparameters) \cite{Tarres2023How2SignSLT}.

\subsubsection{Feature Extraction \& SLT Inference}
We decode generated videos at the dataset’s evaluation FPS and spatial resolution,
then extract \textbf{RGB I3D} features with a Kinetics-pretrained backbone \cite{Carreira_2017_CVPR} .The extracted I3D features are fed \emph{as-is} into the How2Sign SLT model \cite{Tarres2023How2SignSLT}, which produces a translation hypothesis $\hat{y}$ for each segment. 

\subsubsection{SLT Evaluation}
We compute BLEU-$n$ ($n{=}1\ldots4$), rBLEU and chrF++ with \textbf{SacreBLEU}
in the SLT evaluation \cite{Papineni2002BLEU, PopovicChrf, Post2018SacreBLEU}. To control for dataset difficulty and model variance,
we report
\[
\Delta\text{BLEU}_4 \;=\; \text{BLEU}_4(\text{generated}) \;-\; \text{BLEU}_4(\text{ground truth})
\]
where the ``real video'' score is obtained by running the same frozen SLT model
on the ground-truth How2Sign test segments \cite{Duarte2021How2Sign}. Thus, $\Delta\text{BLEU}_4$ close to
$0$ indicates minimal degradation in linguistic intelligibility. Additionally, more SLT details and translation comparison is displayed in Appendix(c. SLT Details) due to page limitation.

\section{Conclusion}
\label{sec:conclusion}
We presented \textbf{SignMimic}, a consistency-aware pipeline for sign-language
video mimicking that disentangles \emph{rigid canonicalization} (TNet), \emph{non-rigid
adaptation} (NIF2D), and \emph{pose-aware completion} (Transformer-based Pose MAE)
prior to a diffusion \emph{inference-only} renderer. This factorization injects
geometric and linguistic priors into the conditioning and yields improved shape and
spatio-temporal consistency, stronger identity preservation, and minimal degradation
under SLT back-translation. 
\section{Future Plan}
\label{sec:future}
Future work will enhance temporal consistency and expand applicability. We plan to lift our Masked Autoencoder (MAE) into the feature space of the diffusion renderer (e.g., SVD), enabling end-to-end training to reduce temporal artifacts like "breathing". Concurrently, we will scale our method to more diverse and in-the-wild sign language corpora, targeting real-world applications such as newsroom deployment which demand near-real-time performance and robustness. Addressing practical safeguards (e.g., consent, watermarking, bias auditing) for responsible dissemination will remain a key priority.
{
    \small
    \bibliographystyle{ieeenat_fullname}
    \bibliography{main}
}

\maketitlesupplementary
\section*{Appendix A. Training and Inference Details}

This section provides additional implementation details for SignMimic,
including datasets and preprocessing, module architectures, training
hyperparameters, and the exact inference pipeline referenced in
Sec.~4.1.2 of the main paper.

\subsection*{A.1. Datasets and Preprocessing}

Table~\ref{tab:dataset_preproc} summarizes the datasets and basic video
preprocessing settings used in our experiments. All datasets follow the
official train/val/test splits described in Sec.~4.1.1.

\begin{table}[h]
    \centering
    \small
    \begin{tabular}{lcccc}
        \toprule
        Dataset & FPS & Resolution &  Role \\
        \midrule
        ASL 50K         & 30  & $640{\times}360$  & pretrain \& eval \\
        How2Sign (FF)   & 25  & $1280{\times}720$  & eval \& SLT \\
        CSL News        & 30  & $200{\times}170$  & eval \\
        Phoenix-2014    & NA  & $210{\times}260$  & eval \\
        \bottomrule
    \end{tabular}
    \caption{Datasets and basic preprocessing. All videos are decoded at a
    fixed frame rate (FPS) and resized so that the signer occupies the
    central region with sufficient background context. ``FF'' denotes the
    front-facing split.}
    \label{tab:dataset_preproc}
\end{table}

We extract whole-body 2D poses (17 body, 6 feet, 68 face, 42 hands;
133 joints total) with DWPose, and use RTMPose as a fallback detector
for rare failure cases.\footnote{Both detectors are used with their
publicly released checkpoints.} For each driving clip we select a single
reference frame for the target signer, typically the first front-facing
frame with high detector confidence.

For each frame $t$, we denote keypoints and confidences as
$P_t \in \mathbb{R}^{J \times 2}$ and $C_t \in [0,1]^J$. As described in
Sec.~3.2, we perform per-part rooting and scale normalization to obtain
$\hat P_t$, which is used by all downstream modules (TNet, NIF2D and
Pose MAE). For Pose MAE training, we further apply confidence-, temporal-
and spatial masking to simulate realistic occlusions and missingness.

\subsection*{A.2. Module Architectures}

Table~\ref{tab:module_arch} summarizes the main architectural choices
for each component in SignMimic. The design closely follows the
description in Sec.~3, and keeps each module lightweight to enable joint
training on multiple datasets.

\begin{table}[h]
\centering
\small
\begin{tabular}{lcccc}
    \toprule
    Module & Input & Output  \\
    
    \midrule
    TNet (per part) &
    $P_t^{\text{part}}$ &
    $2{\times}2$ affine $A_p$ \\
    
    Hier. Pose Grafting &
    $\hat P_t, \hat P_{\text{ref}}$ &
    $P_t^{\text{canon}}$  \\
    
    Identity encoder $E_{\text{id}}$ &
    $\hat P_{\text{ref}}$ &
    $z_{\text{ref}} \in \mathbb{R}^{d_z}$ \\
    
    NIF2D $f_\theta$ &
    $p^{\text{canon}}_i, z_{\text{ref}}$ &
    $\Delta p_i$ \\
    
    Pose MAE encoder &
    pose tokens (per part) &
    latent sequence  \\
    
    Pose MAE decoder &
    mask tokens + encoder feats &
    reconstructed pose  \\
    
    Diffusion backend &
    $\tilde P_t$ &
    RGB video \\
    \bottomrule
\end{tabular}
\caption{High-level architecture of each module. Depth / hidden sizes
are representative and can be adjusted to the exact configuration used
in the implementation (e.g., $d_z$, $L_e$, $L_d$).}
\label{tab:module_arch}
\end{table}

In more detail, the Hierarchical Pose Grafting module (Sec.~3.3) comprises three expert encoders (body, face, shared hands), where each expert begins with a lightweight TNet that regresses $A_p$, followed by shared $1{\times}1$ convolutions and global pooling to produce a part descriptor. Given normalized driving and reference poses, we encode four parts for each, concatenate the eight descriptors, and decode grafted whole-body keypoints in the canonical space. The identity encoder $E_{\text{id}}$ maps the reference pose to a compact latent $z_{\text{ref}}$ that captures signer-specific morphology, which conditions NIF2D to predict per-joint offsets $\Delta p_i$. Pose MAE uses per-part tokenization, temporal 1D convolutions, and a cross-part Transformer to encode visible joints, while a shallow Transformer decoder reconstructs masked joints under the composite loss described in Table~1 of the main paper. The diffusion networks (MimicMotion or latent video diffusion) are kept frozen and only serve as pose-conditioned renderers.

\subsection*{A.3. Training Hyperparameters}

Table~\ref{tab:train_hparams} lists the main optimization settings for
each trainable module. All models are trained with mixed precision on
modern GPUs (e.g., V100/A100). Numbers here can be adapted to the exact
values used in your implementation.

\begin{table}[h]
    \centering
    \small
    \begin{tabular}{lccc}
    \toprule
     & TNet & NIF2D & Pose MAE \\
    \midrule
    Optimizer      & AdamW          & AdamW          & AdamW          \\
    LR             & $1{\times}10^{-4}$ & $2{\times}10^{-4}$ & $1{\times}10^{-4}$ \\
    Batch          & 128 seq         & 128 clips      & 128 clips      \\
    Epochs / iters & 100 epochs     & 100 epochs      & 80 epochs      \\
    Params (M)     & 0.1M          & 0.3M             & 22M             \\
    \bottomrule
    \end{tabular}
    \caption{Training hyperparameters and parameter counts for SignMimic modules.
    ``Seq'' denotes pose sequences; ``clips'' denotes driving-clip–level batches.
    Learning rates, batch sizes and parameter counts can be updated to match the
    actual implementation.}
    \label{tab:train_hparams}
\end{table}

\begin{table*}[t]
    \centering
    \small
    \begin{tabular}{lccccc}
    \toprule
    Method & FID$_\text{VID}$ $\downarrow$ & FVD $\downarrow$ & ID COS $\uparrow$ & PSNR-TS $\uparrow$ & SSIM-TS $\uparrow$ \\
    \midrule
    SignMimic (ours)            &  \textbf{55.81} &  \textbf{471.37} &  \textbf{0.4451} & \textbf{28.90} & \textbf{0.9634} \\
    MimicMotion (2025)                 & 111.38 & 1309.06 &  0.1523 & 23.75 & 0.9038 \\
    Animate Anyone (2024) & 135.95 & 1499.15 &  0.2681 & 27.49 & 0.9345 \\
    Magic Pose (2024)                  & 163.91 & 1315.78 & 0.1460 & 25.24 & 0.8976 \\
    \bottomrule
    \end{tabular}
\caption{Additional benchmark comparison on ASL 50K. Lower is better for
FID$_\text{VID}$ and FVD; higher is better for ID COS, PSNR-TS, and SSIM-TS. The best result is highlighted in bond font.}
\label{tab:asl_extra_bench}
\end{table*}

\begin{table*}[t]
    \centering
    \small
    \caption{SLT evaluation results (corpus-level). rBLEU and rchrF2++ denote the reduced versions of BLEU and chrF2++.}
    \label{tab:app_slt}
    \begin{tabular}{lccccccc}
        \toprule
        Dataset & BLEU-1 & BLEU-2 & BLEU-3 & BLEU-4 & rBLEU & chrF2++ & rchrF2++ \\
        \midrule
        How2Sign & 57.8 & 26.8 & 18.3 & 9.9 & 8.4 & 40.6 & 20.3 \\
        \bottomrule
    \end{tabular}
\end{table*}

\subsection*{A.4. Inference Details}

At test time, we follow the fixed pipeline outlined in Sec.~4.1.2.
Given a driving video and a single reference frame, we:

\begin{enumerate}
    \item Extract 2D whole-body keypoints and confidences
    $(P_t, C_t)$ with DWPose for all driving frames and for the
    reference frame.
    \item Apply per-part rooting and scale normalization (Eq.~(1)) to
    obtain $\hat P_t$ and $\hat P_{\text{ref}}$.
    \item Run the Hierarchical Pose Grafting network with TNet to map
    $(\hat P_t, \hat P_{\text{ref}})$ into the canonical space,
    yielding $P^{\text{canon}}_t$.
    \item Compute the identity embedding
    $z_{\text{ref}} = E_{\text{id}}(\hat P_{\text{ref}})$ and apply
    NIF2D to obtain retargeted poses $P^{\text{retarget}}_t$.
    \item Construct pose-aware masks from confidence thresholds and
    visibility cues, and feed $(P^{\text{retarget}}_t, M_t)$ into the
    Pose MAE to inpaint occluded joints and smooth temporal dynamics,
    producing $\tilde P_t$.
    \item De-normalize $\tilde P_t$ back to image coordinates and pass
    the sequence to the frozen MimicMotion (or SVD-style) generator to
    synthesize the final video of the target signer.
\end{enumerate}

This procedure is deterministic given a driving video, reference frame,
and detector outputs, and is used for all quantitative and qualitative
results reported in the main paper.

\section*{Appendix B. Interpolation Experiment}

\begin{table}[h]
\centering
\small
\begin{tabular}{lcccc}
\toprule
Model & FID $\downarrow$ & ID COS $\uparrow$ & PSNR-TS $\uparrow$ & SSIM-TS $\uparrow$ \\
\midrule
\texttt{interp\_0}  & 52.7979 & 0.3409 & 28.5928 & 0.9643 \\
\texttt{interp\_3}  & 52.5100 & 0.3425 & 28.7500 & 0.9651 \\
\texttt{interp\_6}  & 52.5600 & 0.3454 & 28.8600 & 0.9655 \\
\texttt{interp\_10} & 51.7726 & 0.3478 & 29.0488 & 0.9665 \\
\bottomrule
\end{tabular}
\caption{Interpolation experiment on ASL 50K. Increasing interpolation strength
from \texttt{interp\_0} to \texttt{interp\_10} yields small but consistent
improvements in FID, identity preservation, and temporal reconstruction quality.}
\label{tab:interp}
\end{table}

In this experiment we examine whether SignMimic can be used as a sentence-level smoother by interpolating between two word-level clips. Concretely, given a preceding clip and a following clip of the same signer, we treat the intermediate gap as missing poses and feed a short context window around the boundary into our Pose MAE with a high temporal mask ratio. The model predicts a sequence of in-between poses that smoothly bridges the two segments, which are then rendered by the frozen diffusion backend. As shown by quantitative results in Tab. \ref{tab:interp}, we evaluated metrics on different step of this interpolation and witnessed better result with larger steps, especially on video continuity.

\section*{Appendix C. SLT Details}
\label{sec:app_slt}

For sign language translation (SLT), we report corpus-level BLEU and chrF2++ scores
following the standard sacreBLEU setup. As shown in Tab.~\ref{tab:app_slt},
our current SLT model achieves modest n-gram overlap, reflecting the difficulty
of the task and the relatively small training corpus. In our work, SLT serves
primarily as an auxiliary signal to verify that better mimicking quality
translates into improved downstream recognition and translation performance.

\section*{Appendix D. Extra Benchmark Experiment}

As shown in Tab.~\ref{tab:asl_extra_bench}, we further compare SignMimic against recent video-based sign mimicking baselines on ASL 50K, including MimicMotion (2025 SOTA) and two 2024 SOTA methods, Moore Thread/Animate Anyone and Magic Pose. SignMimic consistently outperforms all three baselines across all metrics, achieving substantially lower FID$_\text{VID}$ (55.8 vs.\ 111.4 for MimicMotion) and FVD (471 vs.\ 1309), while also improving identity preservation (ID COS), as well as temporal reconstruction quality (PSNR-TS and SSIM-TS). This indicates that our pose-aware conditioning pipeline not only reduces visual artifacts but also better maintains signer identity and temporal coherence than prior SOTA systems.

\end{document}